\def\year{2021}\relax
\documentclass[letterpaper]{article} 
\usepackage{aaai21}  
\usepackage{times}  
\usepackage{helvet} 
\usepackage{courier}  
\usepackage[hyphens]{url}  
\usepackage{graphicx} 
\usepackage{natbib}  
\usepackage{caption} 
\usepackage{amsmath}
\usepackage{amsfonts}
\usepackage{booktabs}
\usepackage{subcaption}
\usepackage[group-separator={,}]{siunitx}

\newenvironment{salign}{\begin{equation}\begin{aligned}}{\end{aligned}\end{equation}}
\newenvironment{proof}{{\noindent\it Proof.}\quad}{\hfill $\square$\par}
\def\VAR#1{{\rm Var}[#1]}
\def\E#1{{\rm E}[#1]}
\def\SE#1{{\rm E}^2[#1]}
\title{Delving into Variance Transmission and Normalization: Shift of Average Gradient Makes the Network Collapse}
\author{
	Yuxiang Liu\textsuperscript{\rm 1},
	Jidong Ge\textsuperscript{\rm 1},
	Chuanyi Li\textsuperscript{\rm 1} and 
	Jie Gui \textsuperscript{\rm 2} \\
}
\affiliations{
    \textsuperscript{\rm 1} State Key Laboratory for Novel Software Technology, Nanjing University\\
    \textsuperscript{\rm 2} School of Cyber Science and Engineering, Southeast University\\

    yxliu@smail.nju.edu.cn, \{gjd, lcy\}@nju.edu.cn, guijie@seu.edu.cn
}

\begin{document}

\maketitle

\begin{abstract}
	Normalization operations are essential for state-of-the-art neural networks and enable us to train a network from scratch with a large learning rate (LR). We attempt to explain the real effect of Batch Normalization (BN) from the perspective of variance transmission by investigating the relationship between BN and Weights Normalization (WN). In this work, we demonstrate that the problem of the shift of the average gradient will amplify the variance of every convolutional (conv) layer. We propose Parametric Weights Standardization (PWS), a fast and robust to mini-batch size module used for conv filters, to solve the shift of the average gradient. PWS can provide the speed-up of BN. Besides, it has less computation and does not change the output of a conv layer. PWS enables the network to converge fast without normalizing the outputs. This result enhances the persuasiveness of the shift of the average gradient and explains why BN works from the perspective of variance transmission. The code and appendix will be made available on https://github.com/lyxzzz/PWSConv.
	
\end{abstract}

\section{Introduction}
We have witnessed the growth of deep learning in computer vision. With the maturity of theories  for convolutional neural networks (CNNs)~\cite{DBLP:journals/neco/LeCunBDHHHJ89,krizhevsky2012imagenet}, several research directions, such as image classification~\cite{DBLP:journals/corr/SimonyanZ14a,DBLP:conf/cvpr/HeZRS16,DBLP:conf/cvpr/SzegedyVISW16,DBLP:conf/cvpr/HuangLMW17} and object detection~\cite{DBLP:conf/cvpr/RedmonDGF16,DBLP:conf/nips/RenHGS15,DBLP:conf/eccv/LiuAESRFB16}, can be explored in depth.

Batch Normalization (BN)~\cite{DBLP:conf/icml/IoffeS15}, benefiting from its ability to accelerate training speed and reduce the impact of complex initialization and variance transmission~\cite{glorot2010understanding,DBLP:conf/iccv/HeZRS15}, has been widely used in most of the state-of-the-art networks. With the help of normalization, the setting of learning rate (LR) is no longer scrupulous. 
Generally, the network (without BN) will collapse rapidly if a large LR is set for training. Intuitively, this is attributed to a large update caused by inappropriate LR. By contrast, the appearance of BN alleviates this phenomenon. Moreover, BN accelerates the convergence only by modifying the mean and variance of the output layer, which is puzzling.

Despite normalization's pervasiveness, the real reason for why they are useful still needs to be studied. However, BN is not omnipotent. It tends to use large batch size and will crash by inaccurate batch statistics estimation when we use a miniature batch size. In ~\cite{DBLP:conf/nips/SanturkarTIM18, bjorck2018understanding}, BN was thought to make the optimization landscape significantly smoother. Many variants~\cite{DBLP:journals/corr/BaKH16,ulyanov2016instance,DBLP:conf/eccv/WuH18,DBLP:conf/iclr/LuoRPZL19} of BN are proposed. Nonetheless, they may not replace BN entirely because of time cost, performance, and other factors. Inspired by those methods of conducting normalization on outputs, weight normalization (WN)~\cite{DBLP:journals/corr/SalimansK16} and other variants~\cite{DBLP:journals/corr/abs-1903-10520,DBLP:conf/iccv/HuangLLLT17,DBLP:journals/corr/abs-1709-06079} that conduct normalization on filters' weights are proposed. 
Those methods which normalize the filters usually cooperate with BN to speed up convergence. Many of them may not work if the scale of the output is out of control. Although BN and other methods are effective, changing the outputs may block the variance transmission and lose some information. To better understand why normalizing the outputs is beneficial, we intend to analyze BN from the perspective of variance transmission. We attempt to find a method, which is equivalent to BN and simultaneously robust to mini-batch size. In this way, we may find out why BN works to some extent.

In this paper, we study the variance transmission to find out why BN can speed up convergence and keep the network stable when we use a large LR. We point out that the shift of the average gradient, which is reduced by BN and WN inadvertently, will hinder network training. Therefore, we present Parametric Weights Standardization (PWS) as an alternative to BN and WN. PWS solves the shift of the average gradient by normalizing outputs. It needs less computation and is robust to mini-batch size. Moreover, PWS does not directly control the output variance. PWS acts the same as normal conv operation. 
We have carried out object detection task in VOC and COCO~\cite{lin2014microsoft} datasets, and image classification task in CIFAR10 and ImageNet~\cite{russakovsky2015imagenet}. All models are trained from scratch~\cite{DBLP:conf/iccv/ShenLLJCX17,DBLP:conf/cvpr/ZhuZWWSBM19} to exploit the power of normalization. We refocus on variance transmission to find the discrepancy among normalization operations. Our contributions are:
\begin{itemize}
	\item Theoretically, we reveal how the shift of the average gradient makes a significant impact on variance transmission. A huge variance will increase the scale of the gradient. Thus the network collapses. 
	\item We propose Parametric Weights Standardization (PWS), a fast and robust to mini-batch size normalization operation, to solve the shift of the average gradient. 
	Even if we use a large LR, PWS can still be trained without modifying the variance of the output layer. 
	The variance transmits naturally. The relationship between BN and PWS may prove why BN is useful.
	\item The results indicate that normalizing outputs is not unique to get better results and faster convergence speed.
\end{itemize}

\section{The Shift of the Average Gradient}
\subsection{Variance Transmission}
To understand the role of the shift of the average gradient, we first pay attention to the variance transmission and how the scale of variance influences on the gradient. For a conv layer, we can get
\begin{equation}
Y_l = X_l \otimes W_l + b_l, \qquad X_{l+1} = f(Y_l).
\end{equation}
Here, $l$ represents the index of a layer. $\otimes$ represents the convolutional operation. $Y$ is the output feature map, which is a $\hat{w}$-by-$\hat{h}$-by-$d$ tensor. $\hat{w}$ and $\hat{h}$ are the spatial width and height of the output feature map. $d$ represents the number of output channels or the number of filters in that conv layer. $X$ is the input feature map, which is a $R^{w \times h \times c}$ tensor. $w$ and $h$ are the spatial width and height of the input feature map. $c$ represents the number of input channels of that conv layer. $W_l$ is a $R^{k \times k \times c \times d}$ tensor. $k$ is the spatial filter size of that layer. We use $n_l = k^2 c$ to denote the number of units in one filter of that conv layer $l$. $b$ is a $d$-by-1 vector of biases. $f$ is the activation function. In our experiments, $f$ function is ReLU~\cite{DBLP:conf/icml/NairH10}. 

Generally, the elements in $W_l$ will be initialized to be mutually independent of one another and have a symmetric distribution with a zero mean. As in~\cite{glorot2010understanding}, we can assume that the elements in $X_l$ are also mutually independent of each other and share the same distribution. The elements in $W_l$ and the elements in $X_l$ are independent of each other. For any tensor $T$, we write ${\rm Var}[T]$ and ${\rm E}[T]$ for the shared scalar variance and mean of all elements in $T$, respectively. $\|T\|$ denotes the square root of the quadratic sum of all elements in $T$. If we assume that $b_l=0$, ${\rm E}[Y_l] = n_l {\rm E}[X_l]{\rm E}[W_l] = 0$ and $Y_l$ has a symmetric distribution around zero impacted by $W_l$. This leads to:
\begin{equation}
\begin{aligned}
{\rm E}[X_{l}^2] = {\rm E}[relu(Y_{l-1})^2] = \frac{1}{2} {\rm Var}[Y_{l-1}], \\
and \quad {\rm Var}[Y_l] = \frac{1}{2} n_l {\rm Var}[W_l] {\rm Var}[Y_{l-1}].
\end{aligned}
\label{equ2}
\end{equation}
For every layer, one suggestion is to initialize ${\rm Var}[W_l]$ as $\frac{2}{n_l}$~\cite{DBLP:conf/iccv/HeZRS15} to transmit variance through ReLU. 

Then we focus on the backward propagation case. We get a estimation from ~\cite{glorot2010understanding,DBLP:conf/iccv/HeZRS15}: 
\begin{equation}
{\rm Var}[\nabla_{X_{l}} L] = (\prod_{i=l}^{m-1} \theta n_{i+1} {\rm Var}[W_i]) {\rm Var}[\nabla_{X_{m}} L],
\end{equation}
where $m$ represents the number of layers in our networks. $\theta$ is 1~\cite{glorot2010understanding} when $f'(0)=1$ and $\theta$ is $\frac{1}{2}$~\cite{DBLP:conf/iccv/HeZRS15} when $f$ is ReLU. We use ReLU as function $f$. The initialization for ${\rm Var}[W_l]$ directly impacts on the gradients for filters in all layers. ${\rm Var}[Y_l]$ and ${\rm Var}[\nabla_{X_{l}} L]$ may both explode due to an unreasonable initialization for $W$. Since $\nabla_{W_{l}} L = X_l \otimes \nabla_{Y_{l}} L$, $\nabla_{W_{l}} L$ will be affected and the network may collapse. 
The influence of variance transmission will become significant if the outputs are not normalized. It deserves to research how to train the network without normalizing the output.

\begin{figure*}[h]
	\centering
	\subcaptionbox{${\rm Var}[W_{o,l} - \Delta W_{o,l}] / {\rm Var}[W_{o,l}]$}[0.4\linewidth]{
		\includegraphics[width=1.\linewidth]{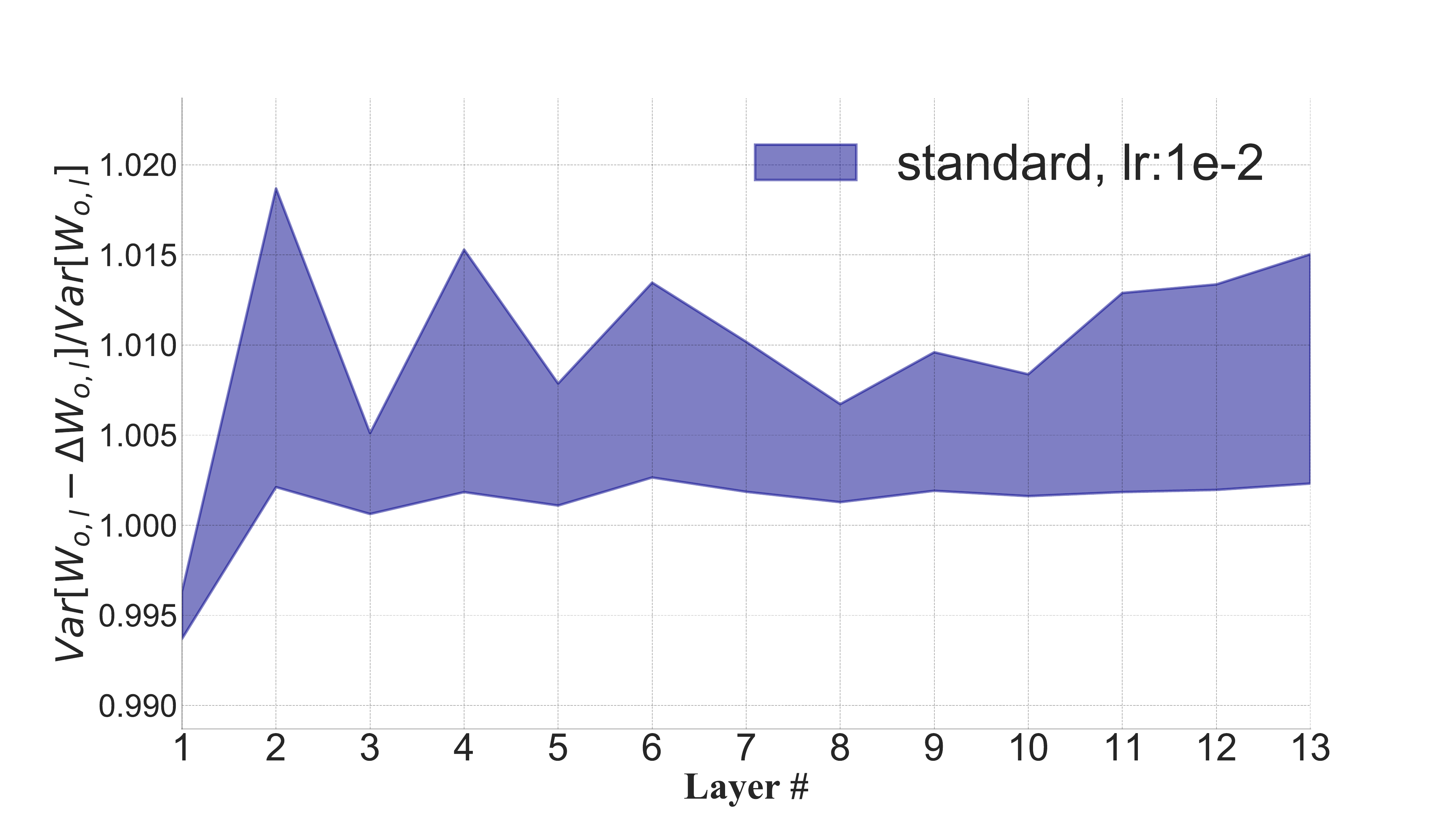}
	}
\qquad
	\subcaptionbox{Average gradients in a layer ($Var[n_l E[\Delta W_{o,l}]]$)}[0.4\linewidth]{
		\includegraphics[width=1.\linewidth]{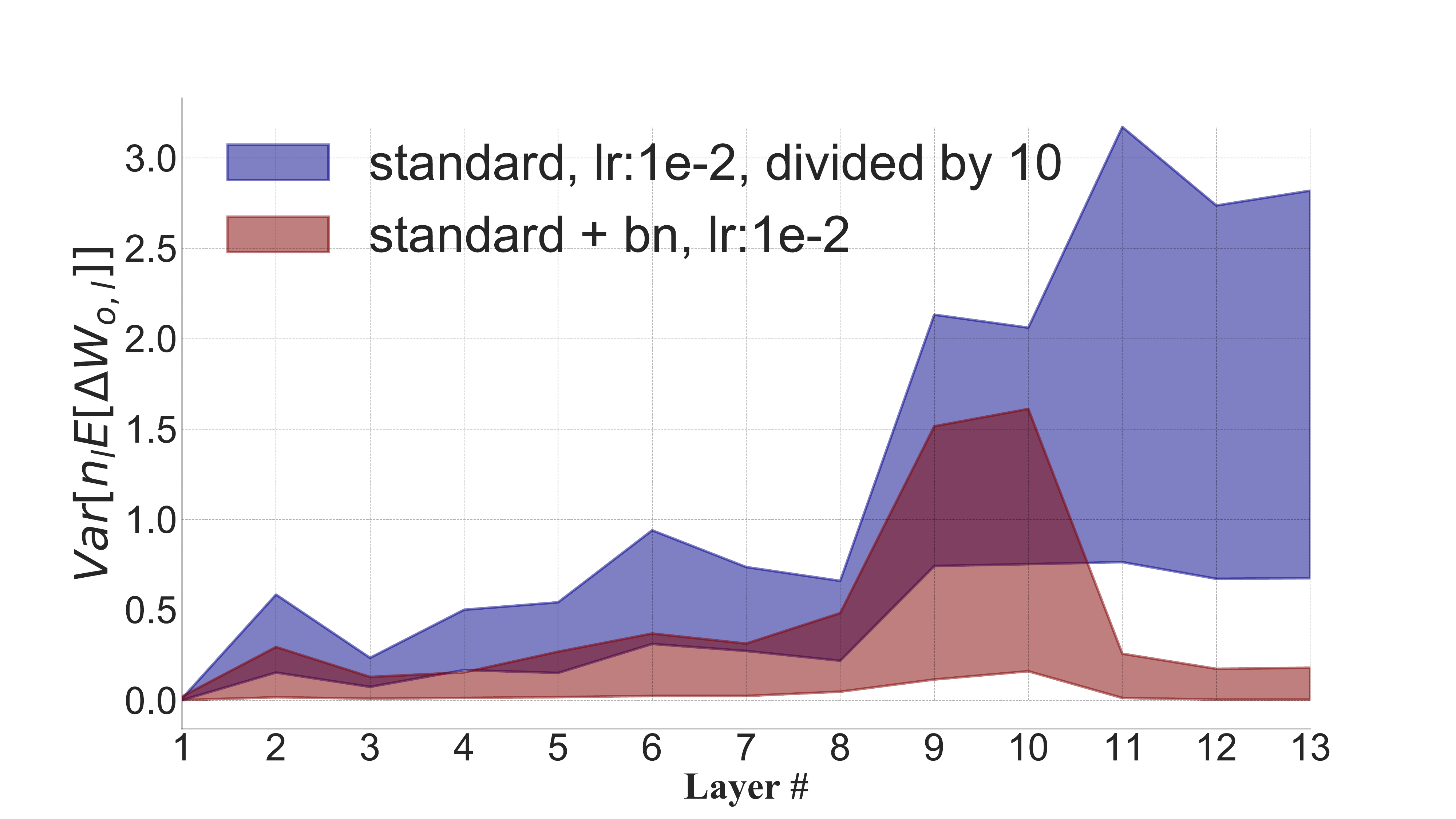}
	}
	\caption{Measuring the gradients for $W_{o,l}$ (Figure a) and the divergence of different filters at the same layer (Figure b). We use $o$ to index a filter in a conv layer. Therefore, $W_l$ is a $k$-by-$k$-by-$c$-by-$d$ tensor and $W_{o,l}$ is a $k$-by-$k$-by-$c$ tensor. $\Delta{W_l}$ can be computed as $LR\times \nabla_{W_l} L$. Figure a shows the influence of a training step on the variance of $W_{o,l}$ using SSD backbone (standard). It is explicit that the variance does not have a huge fluctuation even though the network cannot be trained. Figure b shows the variance of the mean value of different filters at the same layer, represented by ${\rm Var}[n_l {\rm E}[\Delta{W_{o,l}}]]$. The value for the standard backbone is too large, thus we divide it by 10. Figure b may mean that gradients for each filter in a layer have distinct mean value. ${\rm Var}[n_l {\rm E}[\Delta{W_{o,l}}]]$ represents the extent of the shift of the average gradient.}
	\label{tab:figure1}
\end{figure*}

\subsection{Gradient Explosion Comes From the Shift}
Obviously, an inappropriate initialization for weights will cause the gradient explosion. However, the gradient will also explode even though we use a suitable initialization method. We find that a slightly higher LR can also make networks collapse.
Figure~\ref{tab:figure1}(a) displays ${\rm Var}[W_{o,l}-\Delta W_{o,l}]/{\rm Var}[W_{o,l}]$ for every layer. Subscript $o$ is used to index a filter in a layer. $\Delta W_{o,l}$ represents $LR\times \nabla_{W_{o,l}} L$. Here we use SSD~\cite{DBLP:conf/eccv/LiuAESRFB16} + VGG~\cite{DBLP:journals/corr/SimonyanZ14a} as the object detection model because it is sensitive to LR. If we set LR as 1e-2 and do not use BN after every conv layer, the network will collapse. It seems that using a large LR will not change the variance a lot (Figure~\ref{tab:figure1}(a)), but prompts the filter to have a large shift (Figure~\ref{tab:figure1}(b)). 
The changes of variance for different filters are insignificant (Figure~\ref{tab:figure1}(a)). Nevertheless, the network indeed collapses. 

The collapse drives us to define a fine-grained variance propagation formula.
Ignoring the bias, we can get $Y_{o,l} = X_l \otimes W_{o,l}$ and ${\rm E}[Y_{o,l}] = n_l {\rm E}[W_{o,l}] {\rm E}[X_l]$ as mentioned before. For a conv layer, the number of output channels is equal to the number of filters. We use $N_l=d$ to denote the number of filters in layer $l$. According to Figure~\ref{tab:figure1}(b), we know ${\rm E}[\Delta W_{o,l}]$ for different filters in the same layer are distinct, especially for networks without BN. This shift has a great impact on variance transmission. $Y_{o,l}$ is a part of $Y_{l}$, where $Y_l$ is a $\hat{w}$-by-$\hat{h}$-by-$d$ tensor and $Y_{o,l}$ is a $\hat{w}$-by-$\hat{h}$ matrix. We can use ${\rm Var}[Y_{o,l}]$ and ${\rm E}[Y_{o,l}]$ to express ${\rm Var}[Y_l]$:
\begin{equation}
{\rm Var}[Y_{o,l}] = \frac{1}{2} n_l {\rm Var}[W_{o,l}] {\rm Var}[Y_{l-1}],
\label{equ4}
\end{equation}
\begin{equation}
\begin{aligned}
{\rm Var}[Y_l] 
=& {\rm Var}[Y_{l-1}] \frac{\sum_{o} \frac{1}{2} n_l {\rm Var}[W_{o,l}]}{N_l} \\
&+ {\rm Var}[n_l {\rm E}[W_{o,l}] {\rm E}[X_l]].
\end{aligned}
\label{equ5}
\end{equation}
A detailed proof is shown in Appendix. Some researches attribute the variance explosion to the change of input distribution and the amplification of the parameters $W_{o,l}$. While Figure~\ref{tab:figure1}(a) shows that ${\rm Var}[W_{o,l}]$ are stable and not amplified too much, even in a collapsed network corresponding to Figure~\ref{tab:figure1}(a). We assume that parameters are initialized reasonably (such as $n_l {\rm Var}[W_{o,l}] = 2$ when ReLU is the activation function) and $n_l {\rm Var}[W_{o,l}]$ are relatively stable at the beginning of the training. $\E{X_l}$ is independent of subscript $o$ and can be regarded as a constant in $\VAR{n_l \E{W_{o,l}} \E{X_l}}$.
Thus we can get: 
\begin{equation}
{\rm Var}[Y_l] \approx {\rm Var}[Y_{l-1}] + {\rm E}^2[X_l] {\rm Var}[n_l {\rm E}[W_{o,l}]].
\label{equ6}
\end{equation}
Generally, variance is transmitted through $\frac{\sum_{o} \frac{1}{2} n_l {\rm Var}[W_{o,l}]}{N_l}$. ${\rm Var}[{\rm E}[W_{o,l}]] \approx 0$ when we initialize $W_l$. Thus ${\rm Var}[Y_l]$ is in normal range transmitted by ${\rm Var}[Y_{l-1}]$. Everything has changed when ${\rm Var}[n_l{\rm E}[W_{o,l}]]$ increases. It is worth mentioning that the shifts of ${\rm E}[W_{o,l}]$ are all coming from gradients because we initialize ${\rm E}[W_{o,l}]$ to 0. Thus the curves in Figure~\ref{tab:figure1}(b) imply an excessive shift on ${\rm E}[W_{o,l}]$. Even though ${\rm Var}[n_l{\rm E}[W_{o,l}]]$ has been divided by 10 for readability, the maximum of this value is up to 5 at initial layers. ${\rm E}^2[X_l]$ is less than $\frac{1}{2} {\rm Var}[Y_{l-1}]$, but ${\rm E}^2[X_l]$ will not be too small because the elements in $X_l$ are all greater than or equal to 0. The range of ${\rm E}^2[X_l]$ and ${\rm E}^2[X_l]/{\rm E}[X_l^2]$ for every layer is shown in Appendix. The increase in ${\rm Var}[Y_l]$ may impact on ${\rm E}[X_{l+1}]$ and is transmitted to latter layers. At last few layers, ${\rm Var}[n_l{\rm E}[W_{o,l}]]$ is theatrically up to 30. Those shifts in different filters are continuously multiplied. Normal transmission may be destroyed by shift of the average gradient.

\subsection{BN's Effect on the Shift of the Average Gradient}
\label{sec:sec2.2}
First, we discuss about how BN reduces the shift of the average gradient. Formally, we can define a simplified BN of a certain channel of a layer $l$ by formula $\hat{Y_o} = \frac{Y_o - \mu_o}{\sigma_o}$, where  $\mu_o = {\rm E}[Y_o]$ and $\sigma_o^2 = {\rm Var}[Y_o] + \epsilon$. Subscript $o$ is used to index an output channel. $\epsilon$ is a small constant. $Y_o$ and $\hat{Y_o}$ is an $N$-by-$w$-by-$h$ tensor. $w$ and $h$ are the spatial width and height of the output feature map. $N$ is the batch size. We use $n=N\times w\times h$ to denote the number of elements in a channel of that output feature map. 
Here we define $\langle U, V\rangle = vec(U)^T vec(V)$, where $vec$ represents flattening a tensor in row major order. Additionally, $\langle U, V\rangle = U \times sum(V)$ if $U$ is a scalar. $sum(V)$ means the sum of all elements in $V$.
We can get:
\begin{equation}
\textstyle \|\nabla_{Y_o} L\|^2 = \frac{1}{\sigma_o^2} (\|\nabla_{\hat{Y}_o} L\|^2 - \frac{\langle 1, \nabla_{\hat Y_{o}} L \rangle^2 + \langle \nabla_{\hat Y_{o}} L, \hat Y_{o} \rangle ^2}{n}),
\end{equation}
\begin{equation}
{\rm E}[\nabla_{Y_o} L] = 0.
\end{equation}
Detailed proof is shown in Appendix. $\|\nabla_{Y_o} L\|$ will be restricted by the $\sigma_o$ compared with $\|\nabla_{\hat Y_o} L\|$. Thus $\sigma_o$ impacts on the gradients for $W_o$. Moreover, ${\rm E}[\nabla_{Y_o} L] = 0$ reduce the shift on the weights when backward propagation. As Figure~\ref{tab:figure1}(b) has shown, the shift for filters with BN is reduced.

For every channel $Y_{o}$ in the output feature map, which is computed by $W_{o}$, BN will make it zero-mean. Therefore, the influence of the shift of the average gradient will be eliminated because BN's zero-mean operation will let ${\rm Var}[{\rm E}[\hat Y_{o,l}]]$ be zero, although it changes the output information. We speculate that we can get an analogous result without scaling if we keep a robust variance transmission. For Layer Normalization (LN)~\cite{DBLP:journals/corr/BaKH16}, Instance Normalization (IN)~\cite{ulyanov2016instance}, Group Normalization (GN)~\cite{DBLP:conf/eccv/WuH18}, and Switchable Normalization (SN)~\cite{DBLP:conf/iclr/LuoRPZL19}, the zero-mean operation of those methods also reduces the shift of the average gradients. 

\subsection{WN's Effect on Variance Transmission}
\label{sec:sec2.3}

Instead of normalizing the outputs, WN divides the filter weights by its Frobenius norm. Formally, we can define a simplified WN of a filter $W$ in a conv layer by $\hat{W} = \frac{g}{\|W\|} W$, where $g$ is a scalar and $W$ is a $k$-by-$k$-by-$c$ tensor. $c$ represents the number of input channels of that conv layer. $k$ is the spatial filter size of that layer. We also use $n_l = k^2 c$ to denote the number of units in a filter of that conv layer $l$. For a conv operation, WN use $\hat{W}$ instead of $W$ to compute the final output. If we set $g$ to a small value, such as 1, instead of Data-Dependent Initialization~\cite{DBLP:journals/corr/SalimansK16}, networks with WN can be trained with a large LR. The reason might be the constrained ${\rm Var}[\hat{W}]$. We can get ${\rm Var}[\hat{W}]$:
\begin{equation}
{\rm Var}[\hat{W}] = \frac{g^2 {\rm Var}[W]}{\|W\|^2} = \frac{g^2}{n_l} \frac{{\rm Var}[W]}{{\rm Var}[W]+{\rm E}^2[W]}.
\end{equation}
The elements in $W$ will be initialized to a symmetric distribution with a zero mean. ${\rm E}[W] \approx 0$ is established at the beginning, leading ${\rm Var}[\hat{W}]$ to $\frac{g^2}{n_l}$. Considering (\ref{equ2}), we know that the value of $\frac{1}{2} n_l {\rm Var}[\hat W_l] = \frac{g^2}{2}$ should be constrained. We assume that ${\rm E}[W] \approx 0$ at the beginning. ${\rm E}[\nabla_{W} L]$ will be constrained when we set $g$ as 1 due to:
\begin{equation}
{\rm E}[\nabla_{W} L] = \frac{g}{\|W\|} {\rm E}[\nabla_{\hat{W}} L],
\end{equation}
\begin{equation}
\|\nabla_{W} L\|^2 =\frac{g^2}{\|W\|^2} (\|\nabla_{\hat W} L\|^2 - \frac{\langle \nabla_{\hat W} L, \hat W \rangle^2}{g^2}).
\label{equ12}
\end{equation}
The detailed proof is shown in Appendix. Filters will be initialized so that $n_l {\rm Var}[W] = \|W\|^2 = 2$ as we mentioned above. $\frac{g}{\|W\|} = \frac{g}{\sqrt{2}}$ directly affect the gradients and the shift of the gradients. A suitable $g$ will prevent the gradient from overflow and shift. However, Data-Dependent Initialization sets $g$ to maintain an equivalent variance between the layer's input and output. It may set $g$ to be relatively large, such as $\sqrt{2}$, which amplifies the shifts and becomes unstable.

Moreover, the gradient for filter $W$ will increase due to the weight decay, and finally numeric overflow if LR is large. (\ref{equ12}) seems to be similar to BN's. $\|\nabla_{\hat W} L\|^2$ will subtract an non-zero value and be constrained by $\|W\|^2$. The input's variance is stable and will not descend a lot during the training. However, $\|W\|$ will be reduced due to the training steps and the weight decay. The descent of $\|W\|$ makes the gradients greater. When we use a large LR and maintain an equivalent variance, the network becomes sensitive and easy to collapse. The detailed illustration is shown in Appendix.

\section{Parametric Weights Standardization (PWS)}

To deal with the shift of the average gradient, we focus on how to solve the gradient shift elegantly without normalizing the outputs.
Generally, normalizing the outputs requires more computation than normalizing the filters.
However, many methods of normalizing filters do not take the output variance and the computation into account. Most of them should be used with BN, GN or other methods of normalizing the outputs. These methods increase the burden of the network to get a better result. To become robust to mini-batch size, and become faster and stabler than those methods, we defined a PWS layer as follow:
\begin{equation}
\begin{aligned}
\hat W_{o} = \sqrt{\frac{2}{n_l}} \frac{W_{o} - {\rm E}[W_{o}]}{\sqrt{{\rm Var}[W_{o}] + \gamma}}, \\
and \quad Y_{o} = \alpha_{o} \cdot X \otimes \hat W_{o} + \beta_{o}.
\end{aligned}
\label{equ11}
\end{equation}
Here $W_o$ is a $k$-by-$k$-by-$c$ tensor. $c$ represents the number of input channels of that conv layer. $k$ is the spatial filter size of that layer. Subscript $o$ denotes the index of a filter at that layer. $n_l = k^2c$ denotes the number of units in a filter of that conv layer $l$. $Y_o$ is a channel of the batch of output feature map as defined above. $X$ is a batch of input feature map. $\alpha_o$, $\beta_o$ and $\gamma$ are scalars. $\alpha_o$ and $\beta_o$ are trainable and $\gamma$ is fixed. ${\rm Var}[W_o]$ and ${\rm E}[W_o]$ represents the calculated variance and mean of all elements in $W_o$. It is crucial to multiply $\frac{W_{o} - {\rm E}[W_{o}]}{\sqrt{{\rm Var}[W_{o}] + \gamma}}$ by $\sqrt{2/n_l}$ to maintain a correct variance of $\hat W_o$. 
Most methods that normalize the filters do not adjust the filters' variance and subsequently amplify the outputs. The variance will explode due to (\ref{equ2}). 
It is straightforward that we can eliminate the shift of the average gradient on different filters by letting ${\rm E}[\hat W_{o}] \equiv 0$ due to:
\begin{equation}
{\rm E}[\nabla_{W_o} L] = - \sqrt{\frac{2}{n_l}}\frac{{\rm E}[\hat{W_o}] \langle \nabla_{\hat{W}_o} L, \hat{W}_o \rangle}{n_l \sqrt{{\rm Var}[W_o]+\gamma}}= 0.
\label{equ13}
\end{equation}
The gradient here is similar to BN's. However, this gradient will directly act on $W_o$. Thus the gradient for $W_o$ will not shift. 
Moreover, ${\rm E}[\hat{W}_o]$ and ${\rm E}[Y_o]$ are correlated due to ${\rm E}[Y_o] = n_l {\rm E}[X] {\rm E}[\hat{W}_o]$ at the beginning of the training, indicating that centralizing the filters is equivalent to centralizing the outputs from the perspective of variance transmission. PWS does not make a strong assumption of letting different inputs to have the same mean value of outputs, which is different from IN.
The relationship between BN and PWS may be that they both solve the shift of the average gradient by making ${\rm E}[Y_{o}]$ in different channels equal. This perspective helps us understand why BN works well and why we can use a large LR with BN. Moreover, PWS can be reset to a normal conv layer in the inference stage because PWS does not change the output variance.
\subsection{Trainable Parameter $\alpha$}
Think about BN's case. In the inference stage, BN will apply constant $\sigma_{o}$ and $\mu_{o}$ on the output. Here $\sigma_{o}$ and $\mu_{o}$ are calculated by moving mean during the training steps. $\sigma_{o}$ and $\mu_{o}$ are fixed values when testing. If we let $P_{o} = W_{o} / \sigma_{o}$ and $Q_{o} = \beta_{o} -\mu_{o} / \sigma_{o}$, the complete operation will just seem like a regular convolution operation that $Y_o = X \otimes P_o + Q_o$. However, the variance of $\hat{W_o}$ is fixed if we use PWS without $\alpha$. Thus the output channels will not be attached to a weight and have the same variance. PWS will let the filters in the same layer to have a different weight with $\alpha$. Simultaneously, $\alpha$ will make PWS layer act the same as conv layer.
\begin{figure*}[h]
	\centering
	\subcaptionbox{Variance of $W_o$}[0.42\linewidth]{
		\includegraphics[width=1.0\linewidth]{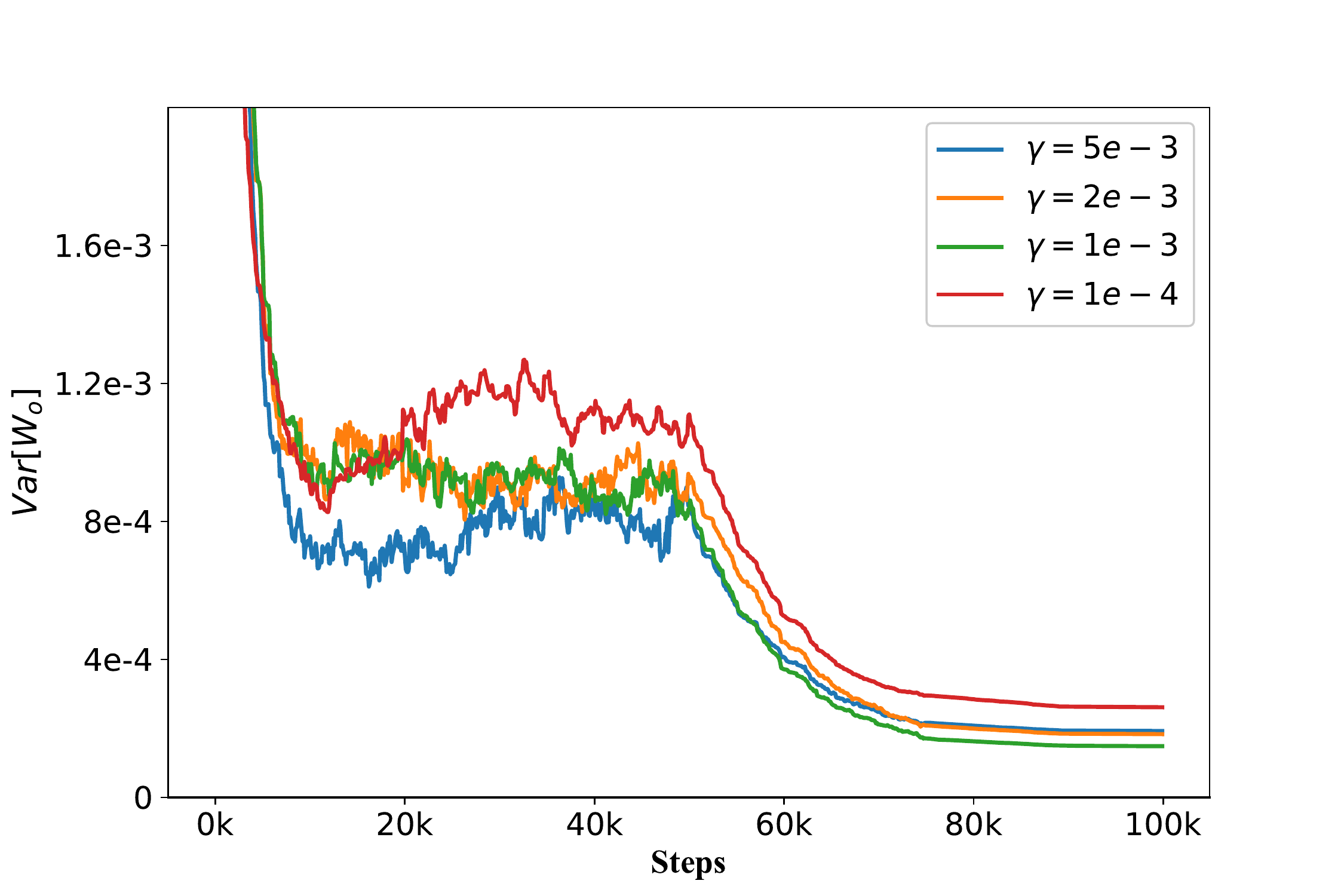}
	}
	\qquad
	\subcaptionbox{$1/\sqrt{Var[W_o]+\gamma}$}[0.38\linewidth]{
		\includegraphics[width=1.0\linewidth]{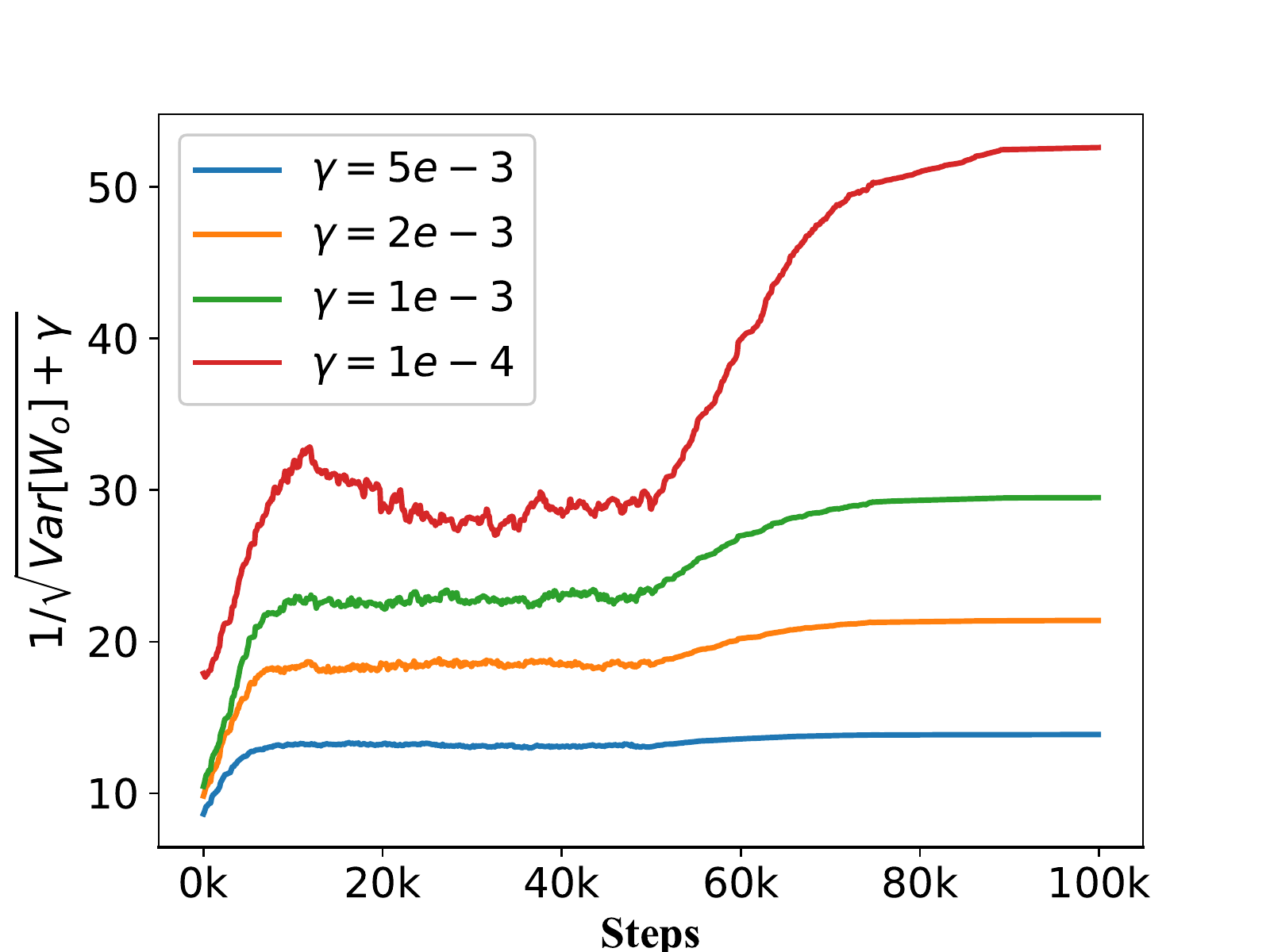}
	}
	\caption{Measuring the minimum of ${\rm Var}[W_o]$ in the third conv layer of the backbone for image classification. The training settings are the same as in the Experiments part except for LR and $\gamma$. Here LR is 5e-2. The curves of different results adopt different $\gamma$. Figure a shows that ${\rm Var}[W_o]$ will decrease when the network is training. ${\rm Var}[W_o]$ decreases faster, especially at the beginning of the training. $\gamma$ will not hinder the training of the network and a proper $\gamma$ will make $\frac{1}{\sqrt{{\rm Var}[W_o] + \gamma}}$ stabler.}
	\label{tab:figure2}
\end{figure*}

\subsection{Parameter $\gamma$}
The difference between normalizing outputs and normalizing filters is that the denominator $\sigma$ in BN is stabler than the denominator in those methods of normalizing filters. $\sqrt{{\rm Var}[W_o]}$ will continuously decrease. Figure~\ref{tab:figure2}(a) shows the variance of $W_o$ during the training. The variance will be reduced due to weight decay. As the training goes on, the gradient for $W_o$ will increase. If we do not use a large LR, the network may benefit from this increase and converge faster. However, the network may collapse when we use a large LR. $\gamma$ in $\sqrt{{\rm Var}[W_o] + \gamma}$ is not acting as $\epsilon$ in BN. As Figure~\ref{tab:figure2}(b) has shown, $\gamma$ can constrain the increase of the reciprocal. Note that with a large LR, even though we set $\gamma$ to 1e-3 (${\rm Var}[W_o]$ is less than 1.6e-3), the network may also converge. A suitable $\gamma$ adjusts the training gradient to an acceptable range when we use a large LR. We will show the results in the Experiment part.

\section{Experiments}
We conduct experiments in object detection and image classification tasks. To prove PWS's usability, we use VGG~\cite{DBLP:journals/corr/SimonyanZ14a} and ResNet~\cite{DBLP:conf/eccv/HeZRS16} as backbones and train all models from scratch with a large LR. It should be mentioned that we remove all BN layers in the backbone to verify our method's effect. 
%
%
For ResNet, we only remove all BN layers. The network structure is not modified.
ReLU is the activation function. We use HE initializer~\cite{DBLP:conf/iccv/HeZRS15} for IN, GN, BN, and SN. We use the same initializer strategy as ~\cite{DBLP:journals/corr/SalimansK16} for WN. For PWS, we find that $\sqrt{2/n_l({\rm Var}[W_o]+\gamma)}$ will impact on $\|\nabla_{W_o} L\|^2$. Suppose we initialize the values of all filters in different layers to follow the same distribution. In that case, the gradients for those filters cannot match the gradients for normal conv operation due to different $n_l$. To match the gradients for normal conv operation, we use HE initializer for PWS rather than set a fixed variance for filters in different layers. $\alpha_o$ is set to 1 and $\beta_{o}$ is set to 0. $\gamma$ is constant for all layers in our experiments. Weight decay for $\alpha_o$ is set as 0. Detailed settings for PWS refer to the configuration in our code repository. Class-wise scores and inference time will also be exhibited in our code repository.

\begin{figure}[tb]
	\centering
	\includegraphics[width=0.65\linewidth]{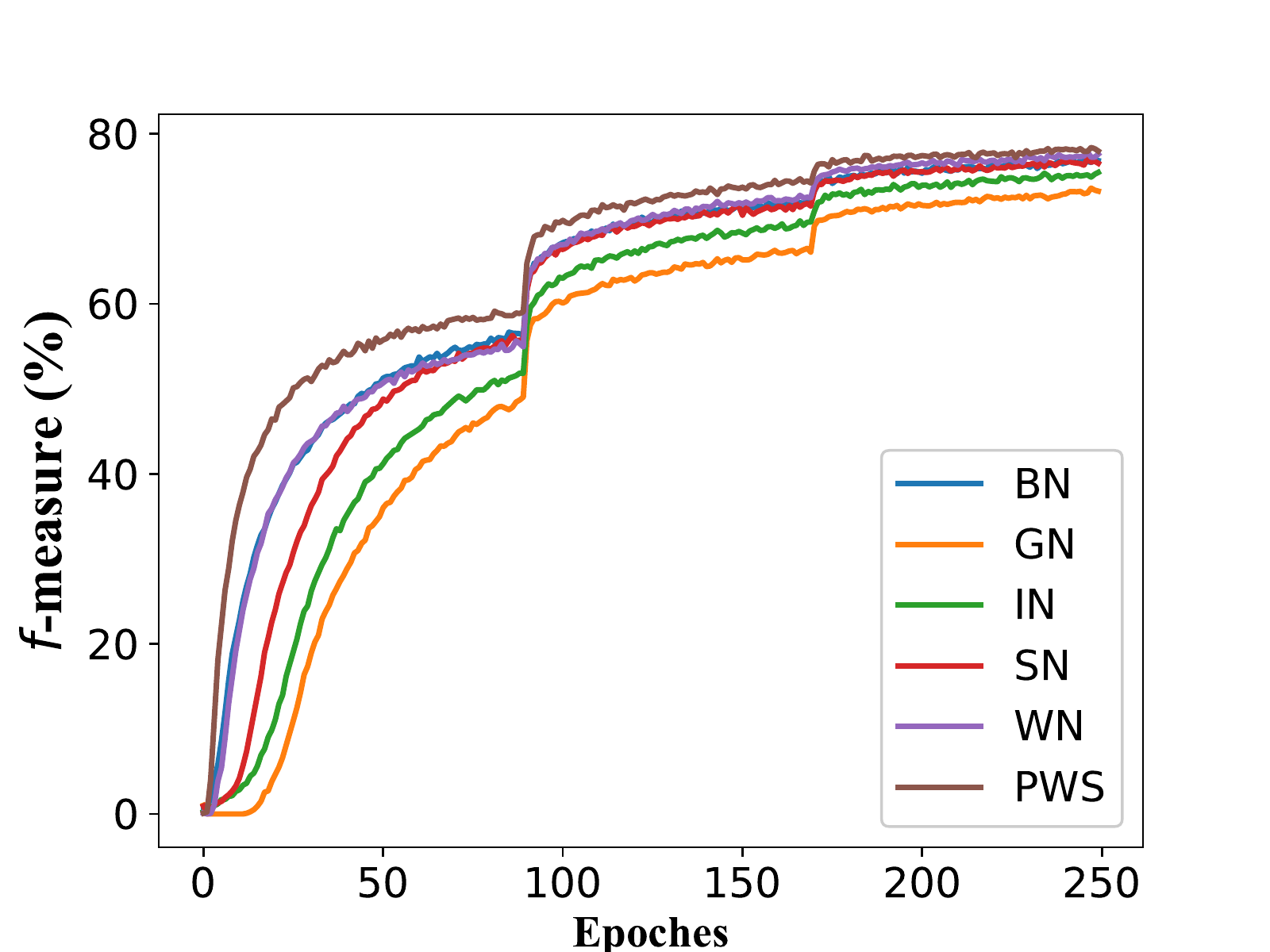}
	\caption{Training conditions of different methods.}
	\label{tab:figure3}
\end{figure}

\subsection{Object Detection in VOC07}
All experiments are conducted on SSD~\cite{DBLP:conf/eccv/LiuAESRFB16} structure. The LR is set to 1e-2 and is multiplied by 0.1 after 90 and 170 epochs. For normal conv operation, it should be mentioned that the network \textbf{cannot be trained} without normalizing outputs when LR is 1e-2. The training finally stops at 240 epochs. We use stochastic gradient descent (SGD) to train the models, where the weight decay is set to 0.0005, and the momentum is set to 0.9. The image size is fixed to 320. We use the same Non-Maximum Suppression (NMS) operation and data augmentation with SSD. All models are trained from scratch in Pascal VOC 07+12 and tested in VOC 07. The train set consists of \num[group-separator={,}]{16551} images, and the test set consists of 4$,$952 images. We do not use any tricks when we train our models and do not use a pre-trained model. $\gamma$ is set to 1e-3 when LR is 1e-2, and 1e-5 when LR is 1e-3.
\subsubsection{Comparison of other normalization methods}
We experiment with different batch sizes to prove that PWS is robust to mini-batch size. Figure~\ref{tab:figure3} shows the training condition with an LR of 1e-2 and a batch size of 16. The $f$-measure is calculated by the predicted object box and the ground truth object box. PWS has a faster convergence speed than other methods. Table~\ref{tab:table1} shows the training speed and the mAP result of different normalization methods in VOC07 test with an LR of 1e-2. These experiments are conducted on a single GPU. Training speed represents number of images that can be trained on an RTX 2080ti per second (batch size$=8$). IN, BN, GN, and SN are methods of normalizing the outputs. IN does not work well as others, and the possible reason is that IN eliminates the difference in different channels for different images. For every image, IN ensures the ${\rm E}[Y_o] \equiv 0$. BN, GN, and SN can work better when training from scratch. However, BN is not robust to mini-batch size (from 76.74 when batch size is 8 to 72.97 when batch size is 2). GN and SN train slower and need more resource. WN and PWS are methods of normalizing the filters. We do not experiment with other methods of normalizing the filters because most of them should be used with BN or GN in computer vision tasks. To make WN feasible to use a large LR, we must manually set $g$ to 1 instead of automatically letting it initialize to ensure the normal spread of variance. Otherwise, the network will collapse at the beginning of the training. However, we found that WN will collapse when using a small batch size. A small batch size leads to more training steps. ${\rm Var}[W_o]$ becomes smaller due to weight decay, which may make the gradients explode.

\begin{table}[t]
	\centering
	\resizebox{0.8\linewidth}{!}{
		\begin{tabular}{@{}ccccc@{}} 
			\toprule 
			& & \multicolumn{3}{c}{Batch Size} \\ 
			\cmidrule(r){3-5}
			Methods & Training speed (FPS) & 2 & 4 & 8\\ 
			\midrule 
			IN & 35.34 & 71.43 & 72.69 & 72.97\\
			BN & 45.66 & 72.97 & 75.18 & 76.74\\
			GN & 29.26 & 75.08 & 75.26 & 75.13\\
			SN & 28.56 & 76.29 & 76.53 & 76.86\\
			\midrule 
			WN $(g=1)$& 52.11 & NaN & 76.70 & 77.06\\
			PWS & 52.72& \textbf{76.93} & \textbf{77.11} & \textbf{77.66}\\ 
			\bottomrule 
		\end{tabular}
	}
	\caption{Sensitivity to batch size: mAP in VOC07, trained with 8, 4 and 2 images/GPU.}
	\label{tab:table1}	
\end{table}
\begin{table}[t]
	\centering
	\resizebox{1.0\linewidth}{!}{
		\begin{tabular}{ccccccc}
			\toprule
			& & &\multicolumn{3}{c}{Component}  & \\ 
			\cmidrule(r){4-6}
			Methods & LR & Backbone &$\sqrt{2/n_l}$ in (\ref{equ11}) & Trainable $\alpha$& $\gamma $& mAP (\%)\\
			\midrule
			Normal conv & 1e-3 & VGG-16 & & & & 68.68\\
			BN & 1e-3 & VGG-16 & & & & 72.80\\
			SN & 1e-3 & VGG-16 & & & & 71.70\\
			PWS & 1e-3 & VGG-16 & \checkmark & \checkmark & 1e-5 & \textbf{72.99}\\
			PWS & 1e-3 & VGG-16 & \checkmark & \checkmark & 1e-3 & 70.32\\
			\midrule
			BN & 1e-2 & ResNet-50 & & & & 75.83\\
			PWS & 1e-2 & ResNet-50 & \checkmark & \checkmark & 1e-3 & \textbf{77.19}\\
			\midrule
			BN & 1e-2 & VGG-16 & & & & 77.00\\
			PWS & 1e-2 & VGG-16 & & & 1e-3 & NaN\\
			PWS & 1e-2 & VGG-16 & \checkmark & & 1e-3 & 74.96\\
			PWS & 1e-2 & VGG-16 & \checkmark & \checkmark & 1e-3 & \textbf{77.52}\\
			PWS & 1e-2 & VGG-16 & \checkmark & \checkmark & 1e-4 & 76.50\\
			PWS & 1e-2 & VGG-16 & \checkmark & \checkmark & 1e-5 & NaN\\
			\bottomrule
		\end{tabular}
	}
	\caption{Ablation study in VOC07 test. Batch size is 16. }
	\label{tab:table2}
\end{table}

\subsubsection{Ablation study for PWS}
Table~\ref{tab:table2} shows the ablation study for PWS. First, without $\sqrt{2/n_l}$, the network will directly collapse (the row of the first NaN result) and the possible reason is that ${\rm Var}[W_{l}] = 1$ leads to the constraint $\frac{1}{2} n_l {\rm Var}[W_{l}]$ to be $\frac{n_l}{2}$. 
The row of the first NaN result verifies that networks may directly collapse if ${\rm Var}[W_o]$ closes to 1 and the outputs are not normalized. This result indicates that some methods that normalize the filters may directly collapse without BN because they do not concern the variance.
The filters' variance must be restricted by this constant parameter to transmit a reasonable variance. The trainable $\alpha$, which is used to change the scales of filters, works well, improving the result from 74.96 to 77.52. PWS works well under different LR when the backbone is VGG. $\gamma$, as we conjecture, constraints the gradient when we use a large LR. The variance of $W_o$ for a conv layer with 64 filters whose spatial size is 3 is 0.00347 ($\frac{2}{3 \times 3 \times 64}$). Therefore, setting a large $\gamma$ will decrease the performance when we use a small LR (1e-3). However, we know that the gradients will increase in WN and PWS. 
The last row of Table~\ref{tab:table2} may verify that the decrease of $\|W_o\|$ indeed destroy the training when we use a large LR. Some results have been shown in Figure~\ref{tab:figure2} to illustrate how $\gamma$ affect the reciprocal. When $\gamma$ is 1e-5, normal LR can make PWS train well (better than BN in Table~\ref{tab:table2}), but excessive LR will lead to NaN.
To avoid the network collapse like WN, we should enlarge $\gamma$. For a large LR, the training may not be hindered by a large $\gamma$. With a suitable setting, PWS can work well and stably without any other normalization operations compared with WN (collapses when using a small batch size) and is more robust to mini-batch size than BN (from 72.97 to 76.93 when batch size is 2).

\begin{table}[t]
	\centering
	\resizebox{1.0\linewidth}{!}{
		\begin{tabular}{@{}ccc|ccc@{}} 
			\toprule 
			Structure & Method & LR schedule & $AP^{bbox}$ & $AP^{bbox}_{50}$ & $AP^{bbox}_{75}$\\ 
			\midrule 
			Mask R-CNN\dag & BN & 1x & 38.0 & 58.6 & 41.4\\
			& GN & 1x & 38.2 & 59.2 & 41.1\\
			& PWS & 1x & \textbf{38.9} & \textbf{59.6} & \textbf{42.5}\\
			\midrule 
			SSD & BN & 1x & 25.2 & \textbf{41.3} & 26.4\\
			& PWS & 1x & \textbf{25.3} & 41.2 & \textbf{26.6}\\
			\cmidrule(r){4-6}
			& BN & 2x & 26.6 & 42.7 & 28.1\\
			& PWS & 2x & \textbf{27.2} & \textbf{44.0} & \textbf{28.7}\\
			\bottomrule 
		\end{tabular}
	}
	\caption{Detection result in COCO. \dag  indicates the implementation by \cite{mmdetection}.}
	\label{tab:table3}	
\end{table}

\subsection{Object Detection in COCO}
We experiment with Mask R-CNN~\cite{he2017mask} + FPN~\cite{lin2017feature} and SSD structure for COCO dataset. The settings for SSD are the same as the experiments in VOC07 except for LR schedule. Batch size is 16 for SSD. LR is set to 1e-2. The standard LR schedule (1x) contains 80 epochs. LR is multiplied by 0.1 after 30, 50, and 70 epochs, respectively. 2x LR schedule contains 150 epochs. LR is multiplied by 0.1 after 60, 100, and 140 epochs, respectively. SSD uses VGG-16 as the backbone. The settings for Mask R-CNN are the same as in \cite{mmdetection}. We use ResNet-50 as backbone. Mini-batch size is 4. The standard LR schedule contains 12 epochs. LR is multiplied by 0.1 after 8 and 11 epochs, respectively. For Mask R-CNN we use pretrained model to alleviate the impact of batch size on BN. The train set consists of 118k images, and the test set consists of 5k images. For PWS conv, we remove all BN layers and replace all conv layers in the backbone with PWS layer. $\gamma$ is set to 1e-4. Details are available on our github.

Table~\ref{tab:table3} shows the results in COCO. For Mask R-CNN, BN gets an inferior result to GN and PWS due to the small batch size. PWS can provide a competitive result with BN and GN and be robust to mini-batch size. For SSD, the results for LR schedule 1x and LR schedule 2x imply that PWS can get similar results to BN under normal batch size. Moreover, PWS does not normalize the outputs and needs less computation, which is distinctive and promising.

\subsection{Image Classification in CIFAR10}
\label{sec:sec4.2}
We experiment with PWS for image classification task in CIFAR10. This dataset consists of 60k images, with 50k training examples and 10k test examples. The LR for networks is multiplied by 0.1 after 100, 150, and 180 epochs, respectively. The training finally stops at 200 epochs. We use stochastic gradient descent (SGD), where the weight decay is set to 0.0005, and the momentum is set to 0.9. The batch size is 128. The image size is fixed to 32. $\gamma$ is set to 1e-3 for PWS. We follow the simple data augmentation in~\cite{DBLP:conf/aistats/LeeXGZT15} for training. All models are trained from scratch.

\begin{figure}[tb]
	\centering
	\includegraphics[width=0.65\linewidth]{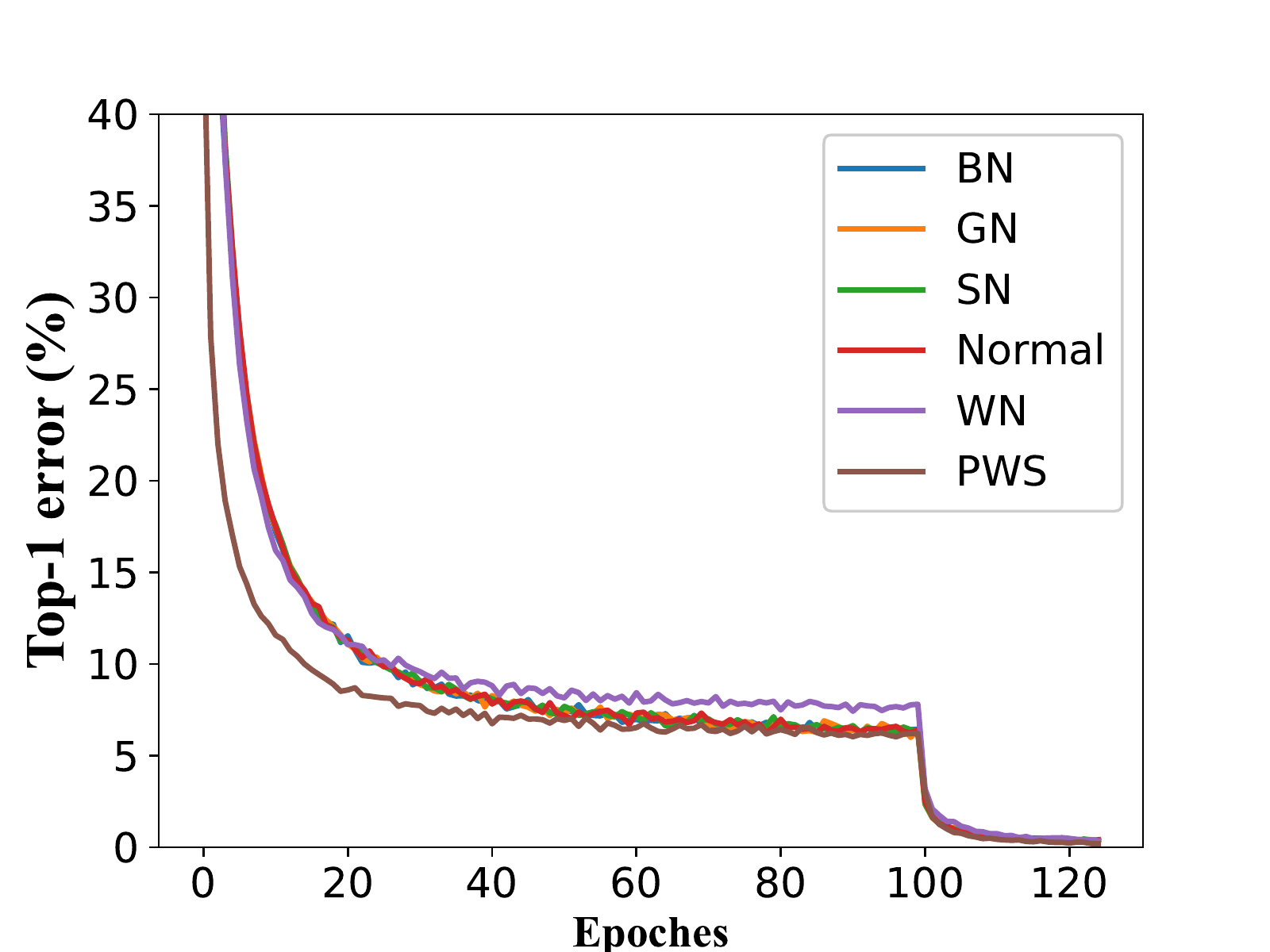}
	\caption{Top-1 error for CIFAR-10 (LR:5e-2).}
	\label{tab:figure4}
\end{figure}

\begin{table}[t]
	\centering
	\resizebox{0.8\linewidth}{!}{	
		\begin{tabular}{@{}ccccc@{}} 
			\toprule
			& \multicolumn{2}{c}{Residual} & \multicolumn{2}{c}{Plain} \\ 
			\cmidrule(r){2-3} \cmidrule(r){4-5}
			Methods & LR: 1e-1 & LR: 5e-2 & LR: 1e-1 & LR: 5e-2 \\ 
			\midrule 
			Normal & 8.82 & 9.11 & 7.28 & 7.40 \\
			BN & \textbf{6.12} & 9.14 & \textbf{5.90} & 7.67 \\
			GN & 6.82 & 9.09 &6.77 &7.54\\
			SN & 6.19 & 8.90 &5.95 &7.34\\
			\midrule 
			WN & 7.18 & 7.41 &6.98 &7.38\\
			PWS & 6.50 & \textbf{6.75} & 6.16 &\textbf{6.30}\\
			\bottomrule 
		\end{tabular}
w	}
	\caption{The classification error for CIFAR-10}
	\label{tab:table4}
\end{table}

We use both plain and residual architectures to verify the generality of PWS. The plain architecture is similar to the ConvPool-CNN-C architecture of~\cite{DBLP:journals/corr/SpringenbergDBR14}, with a modification that abandons all dropout layers. The residual architecture is the same as~\cite{DBLP:conf/cvpr/HeZRS16}. Details about the network architecture will be presented in Appendix. GN does not work well. A possible reason may be that the number of channels in a layer is not large enough. In this experiment, $\gamma$ will not affect a lot because the network is not deep enough. From Table~\ref{tab:table4}, we find BN, GN, and SN will not work better than PWS when LR is small, which may due to the change of output variance. When the LR is set to 1e-1, PWS works not the best. This may be because the gradients for $\alpha$ are too large, leading to some unreasonable updates. Therefore, PWS and WN will both be affected.

\subsection{Image Classification in ImageNet}
\begin{figure}[t]
	\centering
	\includegraphics[width=0.75\linewidth]{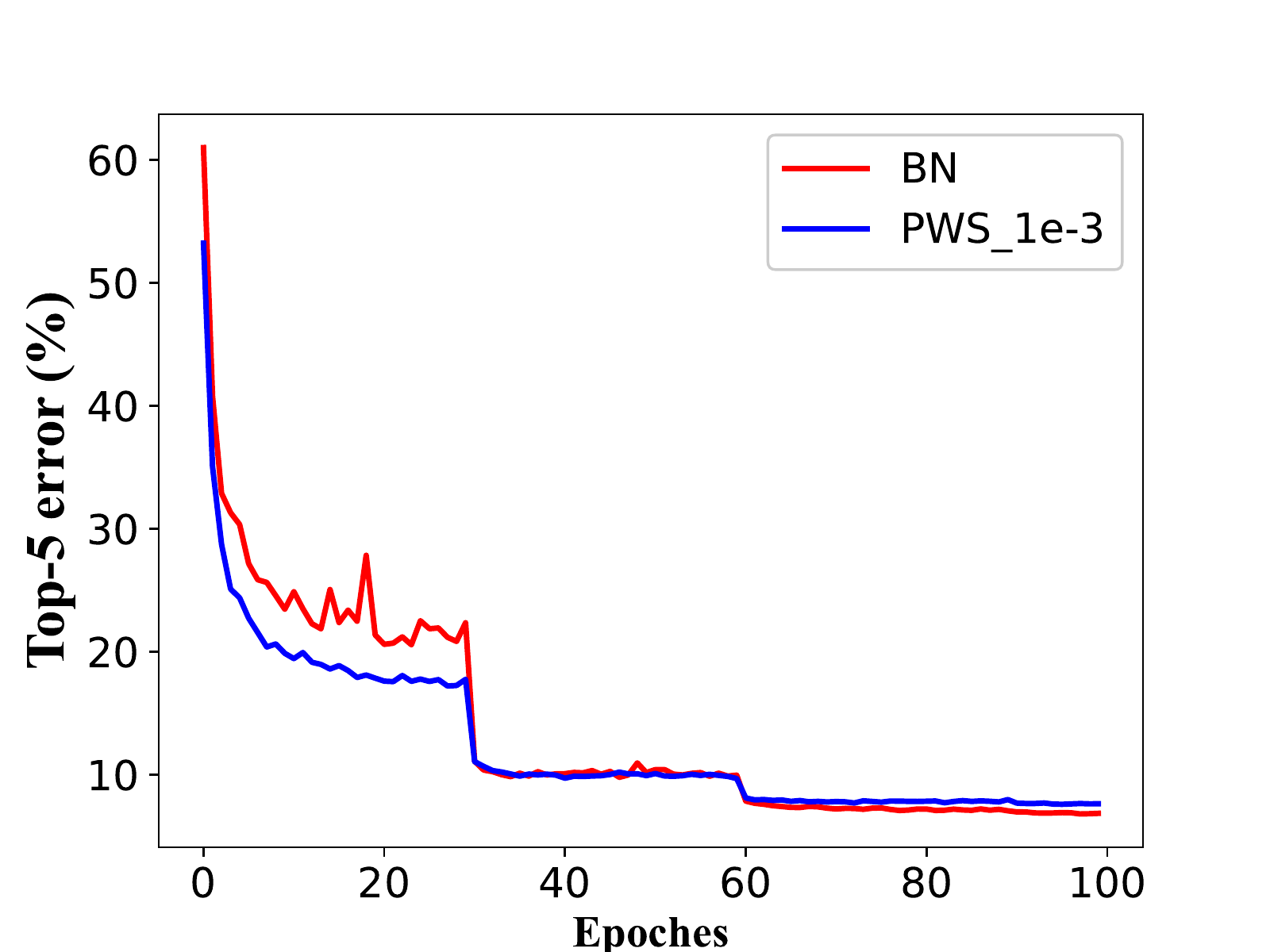}
	\caption{Top-5 error for ImageNet val set.}
	\label{tab:figure5}
\end{figure}

We experiment with PWS for image classification task in ImageNet. We train on the 1.28M training images and evaluate on the 50k validation images. We use ResNet-50 as our backbone. The setting for PWS is the same as we discussed in COCO detection task. To get equivalent outputs and gradients to BN, we use LN to normalize the input and the output of ResNet-50. It should be emphasized that we remove all BN layers in ResNet-50. Batch size is 64. LR is 0.025 at the beginning and multiplied by 0.1 after 30, 60, and 90 epochs. The training finally stops at 100 epochs. We use stochastic gradient descent (SGD), where the weight decay is set to 0.0001, and the momentum is set to 0.9. The image size is fixed to 224. We follow the same data augmentation strategy in ResNet-50. From Table~\ref{tab:table5}, it seems that PWS may not work the best for classification task. However, PWS gets a similar result to GN. On the contrary, WN has degraded performance on ImageNet. PWS is the only method which normalize the filters and get a competitive result.  Moreover, PWS requires less computation (Table~\ref{tab:table1}) and converges faster (Figure~\ref{tab:figure4} and Figure~\ref{tab:figure5}).

\begin{table}[t]
	\centering
	\resizebox{0.8\linewidth}{!}{	
		\begin{tabular}{@{}ccccc|cc@{}} 
			\toprule
			Measurement & BN & GN & LN & IN & WN & PWS\\ 
			
			\midrule 
			val error (\%) & \textbf{23.6} & 24.3 & 25.3 & 28.4 & 28.2 & 24.0\\
			\bottomrule 
		\end{tabular}
	}
	\caption{Comparison of top-1 error rates (\%) of ResNet-50 in the ImageNet validation set.}
	\label{tab:table5}
\end{table}

\section{Conclusion}
PWS is similar to BN. However, it is robust to mini-batch size. The variance transmits naturally with PWS. PWS also has a faster training speed than other normalization operations. The experiment results show that the average gradient migration can affect the network training. Moreover, the results of PWS indicate that normalizing outputs is not the only way to get better results and faster convergence speed.

Nevertheless, it still has problems. We must pay more attention to variance transmission due to the decrease of $\sqrt{{\rm Var}[W_o] + \gamma}$. The network can benefit from a proper $\gamma$, which is different from WN. Therefore, how to find a suitable $\gamma$ automatically remains a problem. Our suggestion is to use a large $\gamma$ when LR is large and use a small $\gamma$ when LR is suitable for normal conv. If your network is not deep, the choice of $\gamma$ will not have a huge impact. One suggestion for the choice of $\gamma$ is to set it as $0.1 \sqrt{n_l/2}$.

In this paper, we theoretically analyze the effect of BN from the perspective of variance propagation. We guess that the shift of the average gradient is a problem, which causes the network to collapse. We propose a fast and robust to mini-batch size method called PWS. It is proved that BN and PWS play the same role in the shift of the average gradient, indicating why BN is beneficial. PWS provides another way to speed up the network fitting. We refocus on variance transmission, which is critical to help us understand how networks work and how to make the network more robust.


\section{Acknowledgments}
This work was supported by National Natural Science Foundation of China (No.61802167, No.61802095), Natural Science Foundation of Jiangsu Province (Grant No.BK20201250), Open Foundation of State key Laboratory of Networking and Switching Technology (Beijing University of Posts and Telecommunications) (SKLNST-2013-1-14), and the Fundamental Research Funds for the Central Universities. Jidong Ge and Jie Gui are the corresponding authors of this paper.

\begin{figure*}[t]
	\centering
	\subcaptionbox{$\SE{X_l}$}[0.4\linewidth]{
		\includegraphics[width=1.\linewidth]{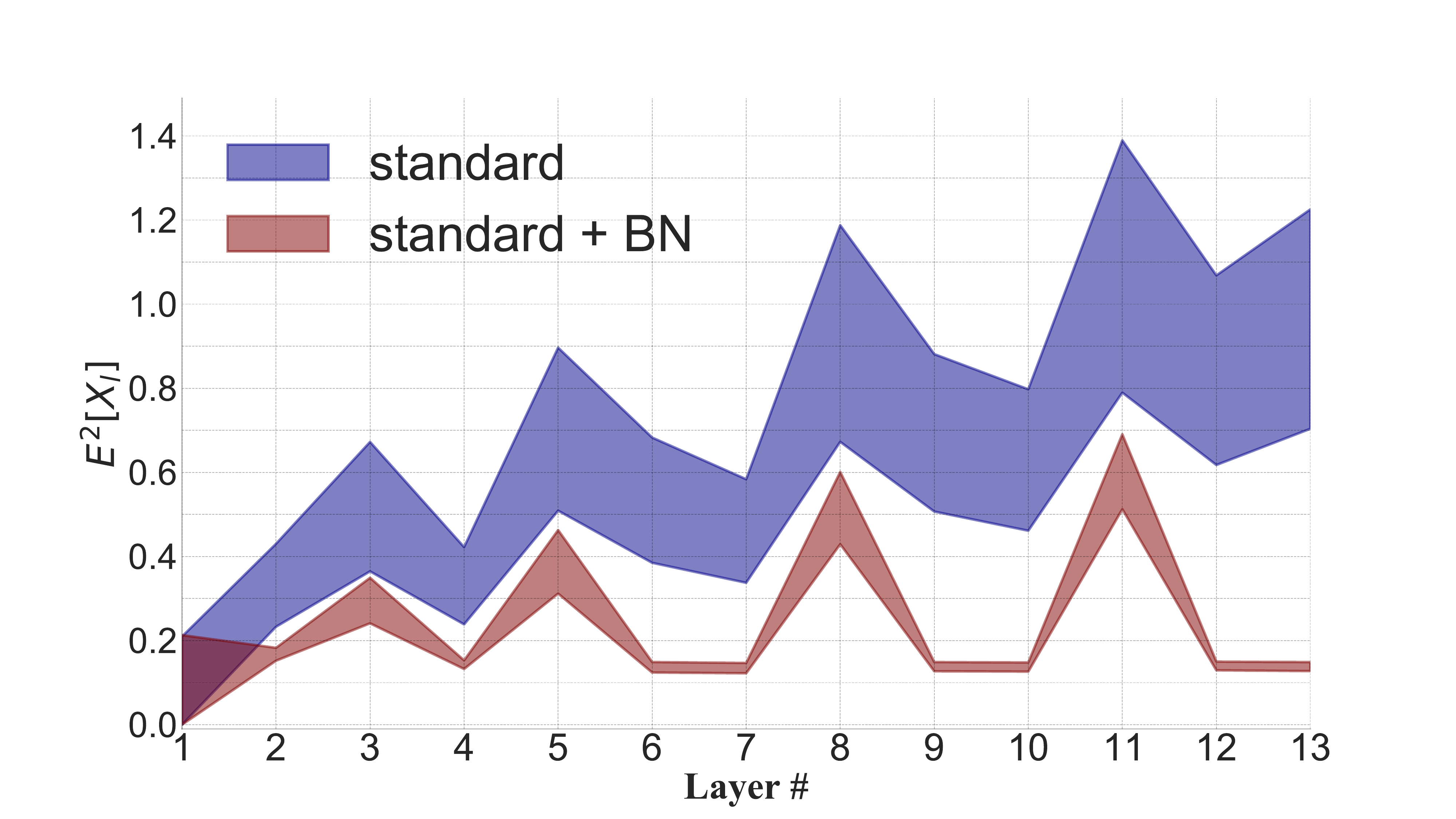}
	}
	\qquad
	\subcaptionbox{$\SE{X_l} / \E{X_l^2}$}[0.4\linewidth]{
		\includegraphics[width=1.\linewidth]{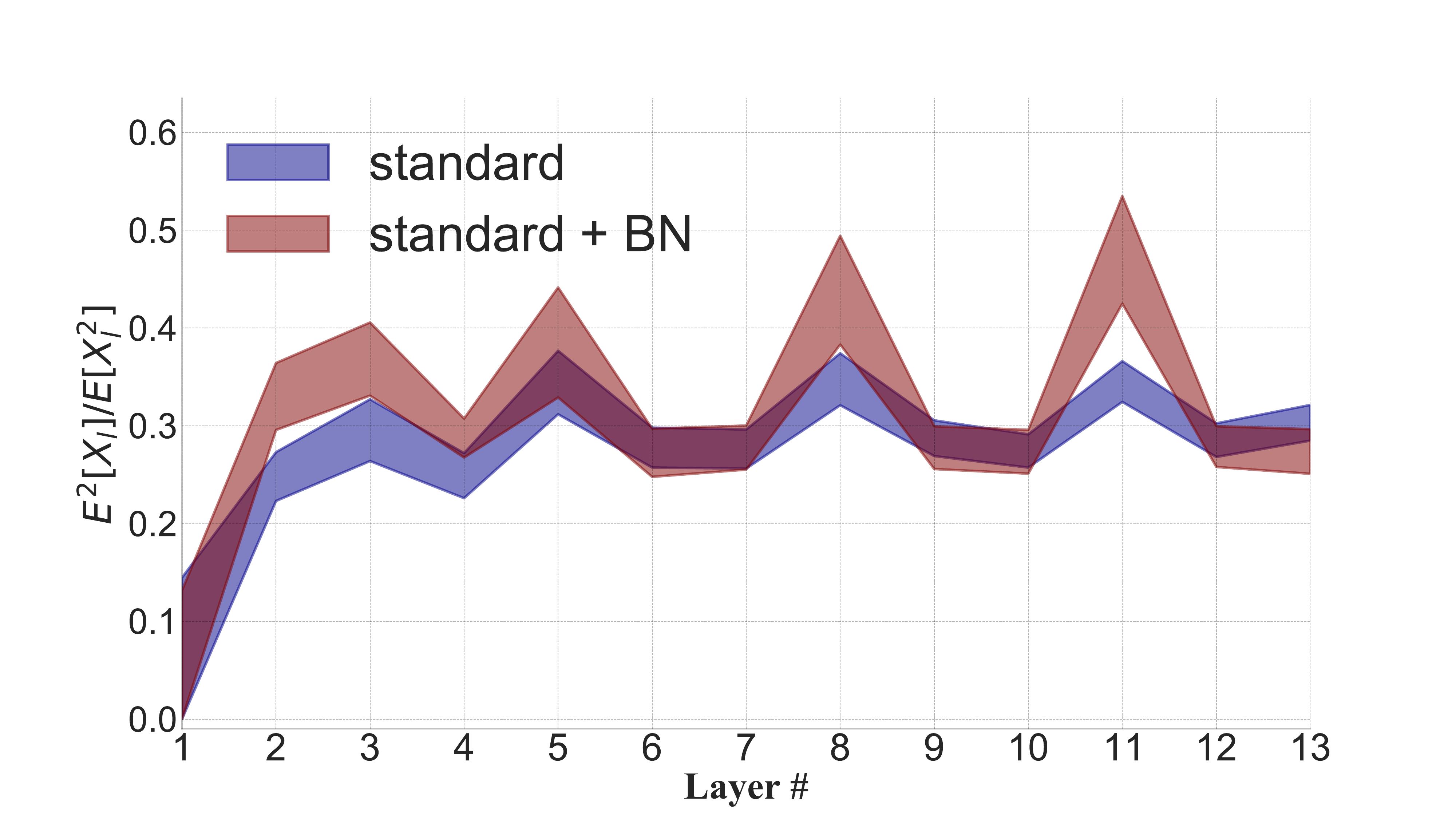}
	}
	
	\caption{The range of ${\rm E}^2[X_l]$ and ${\rm E}^2[X_l]/{\rm E}[X_l^2]$ in different layers using SSD backbone (standard). Here ${\rm E}[X_l]$ does not mean the shared mean of all elements in $X_l$. Here for each input x in $X_l$ at layer $l$, we use $E(x)$ and $E(x^2)$ denotes the expectation of all elements and the expectation of the quadratic of all elements in $x$, respectively. ${\rm E}[X_l]$ denotes the range of $E(x)$ and ${\rm E}[X_l^2]$ denotes the range of $E(x^2)$. We measure these values without updating steps, because the network without BN will directly collapse when we set lr as 1e-2. For every batch, we calculate the forward process and the backward process. Then we calculate what happened when we update the filters' gradients. We do not update the parameters in networks. Therefore, we can collect the result of all training data. The inputs $X_l$ of layers 3, 5, 8, and 11 came from max-pool layer.}
	\label{appendix:1}
\end{figure*}
\section{Appendix}
\subsection{Proof of (\ref{equ6})}
\begin{proof}
	For (\ref{equ6}), we give a detailed proof. We know ${\rm E}[W_{o,l}] = 0$ when we initialize it. Thus we can get:
	\begin{salign}
		\VAR{Y_l} &= \E{Y_l^2} - \SE{Y_l} \\ 
		&= \frac{\sum_o {\rm E}[Y_{o,l}^2]}{N_l} - E^2[Y_l] \\
		&= \frac{\sum_{o} {\rm Var}[Y_{o,l}]}{N_l} + {\rm E}[E^2[Y_{o,l}]] - {\rm E}[{\rm E}[Y_{o,l}]]^2 \\
		&= \frac{\sum_{o} {\rm Var}[Y_{o,l}]}{N_l} + {\rm Var}[{\rm E}[Y_{o,l}]] \\
		&= \frac{\sum_{o} {\rm Var}[Y_{o,l}]}{N_l} + E^2[X_l] {\rm Var}[n_l {\rm E}[W_{o,l}]] \\
		&= {\rm Var}[Y_{l-1}] \frac{\sum_{o} \frac{1}{2} n_l {\rm Var}[W_{o,l}]}{N_l} + E^2[X_l] {\rm Var}[n_l {\rm E}[W_{o,l}]].
	\end{salign}
	
	From Figure~\ref{tab:figure1}(a), we know that the variance of weights are stable. Assume that we set $n_l Var[W_{o,l}]$ as 2 (HE initializer). We may be able to get an approximation:
	\[\sum_{o} \frac{1}{2} n_l \VAR{W_{o,l}} \approx \sum_{o} 1 = N_l.\]
	Then we can get a transition:
	\begin{salign}
		\VAR{Y_l} &\approx \VAR{Y_{l-1}} + \SE{X_l} \VAR{n_l \E{W_{o,l}}} \\
		&= \VAR{Y_{l-1}} (1 + \frac{\SE{X_l}}{\VAR{Y_{l-1}}} \VAR{n_l \E{W_{o,l}}}) \\
		&= \VAR{Y_{l-1}} (1 + \frac{\SE{X_l}}{2 \E{X_l^2}} \VAR{n_l \E{W_{o,l}}})
	\end{salign}
	This completes the proof.\end{proof}

\subsection{The range of 
	\boldmath
	${\rm E}^2[X_l]$
	\unboldmath and 
	\boldmath
	${\rm E}^2[X_l]/{\rm E}[X_l^2]$
	\unboldmath for every layer in the task of object detection}
We know the variance will increase if the mean value for different channels is distinct. This increase does not come from the increase of the ${\rm Var}[W_l]$. We speculate about the reasons for why the network collapses. We find that ${\rm Var}[n_l{\rm E}[\Delta W_{o,l}]]$ will be a large value at the beginning of the training especially for those networks without BN. To make that reason persuasive, we also measure the ${\rm E}[X_l]$ to prove that ${\rm Var}[n_l{\rm E}[W_{o,l}]]$ may have a large impact on variance transmission.

As Figure~\ref{appendix:1} shows, the relationship between $E^2[X_l]$ and ${\rm E}[X_l^2]$ (Figure~\ref{appendix:1} (b)) will not have a large distinction for different networks (comparing standard and BN). BN divides the output variance by $\sigma$. Normally, $\sigma$ is larger than 1. Thus the inputs of the next layer may be smaller when we use BN (Figure~\ref{appendix:1} (a)). $\frac{{\rm E}^2[X_l]}{{\rm E}[X_l^2]}$ may determine the influence of ${\rm Var}[n_l {\rm E}[W_{o,l}]]$.
Assume that $\frac{{\rm E}^2[X_l]}{{\rm E}[X_l^2]}$ is stable at the beginning of the training. It is explicit that the scale of the shifts will have a large impact on the variance transmission.

\bibliography{aaai21}
\end{document}